\documentclass[letterpaper, 10 pt, conference]{ieeeconf}  

\IEEEoverridecommandlockouts                              

\usepackage{graphics} 
\usepackage{epsfig} 
\usepackage{amsmath} 
\usepackage{amssymb}  
\usepackage{hyperref}
\hypersetup{citecolor=black, colorlinks=false, linkcolor=black, pdfborder={0 0 0}}
\usepackage[font=small]{caption}
\usepackage{algorithm,algpseudocode}
\usepackage{booktabs}
\usepackage{subfigure}
\usepackage{todonotes}
\usepackage[per-mode=symbol]{siunitx}
\usepackage{tikz}
\usepackage[utf8]{inputenc}
\usepackage{mathtools}  
\usepackage{xspace}

\DeclareMathOperator*{\argmin}{\text{arg\,min}}
\DeclareMathOperator*{\argmax}{\text{arg\,max}}
\DeclareMathOperator*{\minimize}{\text{minimize}}
\DeclareMathOperator*{\maximize}{\text{maximize}}
\newcommand{\st}{\mbox{\text{subject to}}}
\newcommand{\GaussianDist}[2]{\ensuremath{\operatorname*{\mathcal{N}}\left(#1, #2\right)}}  

\usetikzlibrary{shapes}
\definecolor{blue}{HTML}{1F77B4}
\definecolor{orange}{HTML}{FF7F0E}
\definecolor{green}{HTML}{2CA02E}
\definecolor{red}{HTML}{D62728}
\definecolor{violet}{HTML}{9467BD}
\newcommand{\markerthree}{\raisebox{0.5pt}{\tikz{\node[draw,scale=0.3,regular polygon, regular polygon sides=3,blue,fill=blue,rotate=0](){};}}}
\newcommand{\markerfour}{\raisebox{0.5pt}{\tikz{\node[draw,scale=0.4,regular polygon, regular polygon sides=4,orange,fill=orange](){};}}}
\newcommand{\markerfive}{\raisebox{0.5pt}{\tikz{\node[draw,scale=0.4,regular polygon, regular polygon sides=5,green,fill=green](){};}}}
\newcommand{\markersix}{\raisebox{0.5pt}{\tikz{\node[draw,scale=0.4,regular polygon, regular polygon sides=6,red,fill=red](){};}}}
\newcommand{\markerdiamond}{\raisebox{0pt}{\tikz{\node[draw,scale=0.4,diamond,violet,fill=violet](){};}}}

\makeatletter
\renewcommand{\ALG@beginalgorithmic}{\small}
\makeatother

\title{\LARGE \bf
Computing the racing line using Bayesian optimization
}

\author{Achin Jain$^{1}$ and Manfred Morari$^{1}$
\thanks{$^{1}$Achin Jain and Manfred Morari are with the department of Electrical and Systems Engineering, University of Pennsylvania,
        Philadelphia 19104, PA, USA.
        {\tt\small achinj@seas.upenn.edu, morari@seas.upenn.edu}}
}

\begin{document}

\maketitle
\thispagestyle{empty}
\pagestyle{empty}


\begin{abstract}
	A good racing strategy and in particular the racing line is decisive to winning races in Formula 1, MotoGP, and other forms of motor racing.
	The racing line defines the path followed around a track as well as the optimal speed profile along the path.
	The objective is to minimize lap time by driving the vehicle at the limits of friction and handling capability.
	The solution naturally depends upon the geometry of the track and vehicle dynamics.
	We introduce a novel method to compute the racing line using Bayesian optimization.
	Our approach is fully data-driven and computationally more efficient compared to other methods based on dynamic programming and random search.
	The approach is specifically relevant in autonomous racing where teams can quickly compute the racing line for a new track and then exploit this information in the design of a motion planner and a controller to optimize real-time performance.
\end{abstract}


\section{INTRODUCTION}
\label{S:intro}

The racing line is the single most crucial element of the overall racing strategy in motor racing.
Professional drivers learn from their experience to drive the racing line.
Ahead of any race, the drivers learn the best strategy in a simulator to minimize their lap time.
They practice in the simulator to execute the same strategy and produce best lap times consistently, thus mastering how fast to drive on different parts of the track, when to switch gears, when to start braking as they approach a corner, when to turn in before hitting an apex, when to start accelerating as they exit a corner, etc \cite{Bleacher2019}.
Finally, they get out of the simulator and onto the real track to fine tune their racing strategy to compensate for sim-to-real differences.

Analogously, the algorithms for autonomous racing can exploit the knowledge of a precomputed racing line in the design of a motion planner and a controller, where the goal is to minimize the deviation from the pre-computed racing line.
For example, we can use iterative learning control for lateral path tracking \cite{Kapania2015} or nonlinear model predictive control for motion planning and control \cite{Weiskircher2015}.
Another approach involves using three different controllers, one based on gain scheduling for tracking lateral position, and two proportional controllers for tracking path curvature and velocity \cite{Heilmeier2019}.

The racing line can be either based on a minimum curvature path or a minimum time path.
The former is reasonably close to the latter because it allows the highest cornering speeds at a given maximum lateral acceleration \cite{Heilmeier2019}.
Henceforth, in this paper, we refer to the racing line as the minimum time path.
The resulting optimization is a minimum time control problem which is computationally challenging to solve in general \cite{Athans2013}.
Nonlinear vehicle dynamics make it even harder.
Different ways proposed in the literature to solve this problem include dynamic programming \cite{Beltman2008} which does not scale well, nonlinear optimization solved iteratively \cite{Rosolia2019} which is complex and requires expert domain knowledge to implement and tune, and random population-based search using genetic programming \cite{Vesel2015} which requires tuning and takes a long time to converge.

\begin{figure}[t]
	\centering
	\includegraphics[width=1\columnwidth]{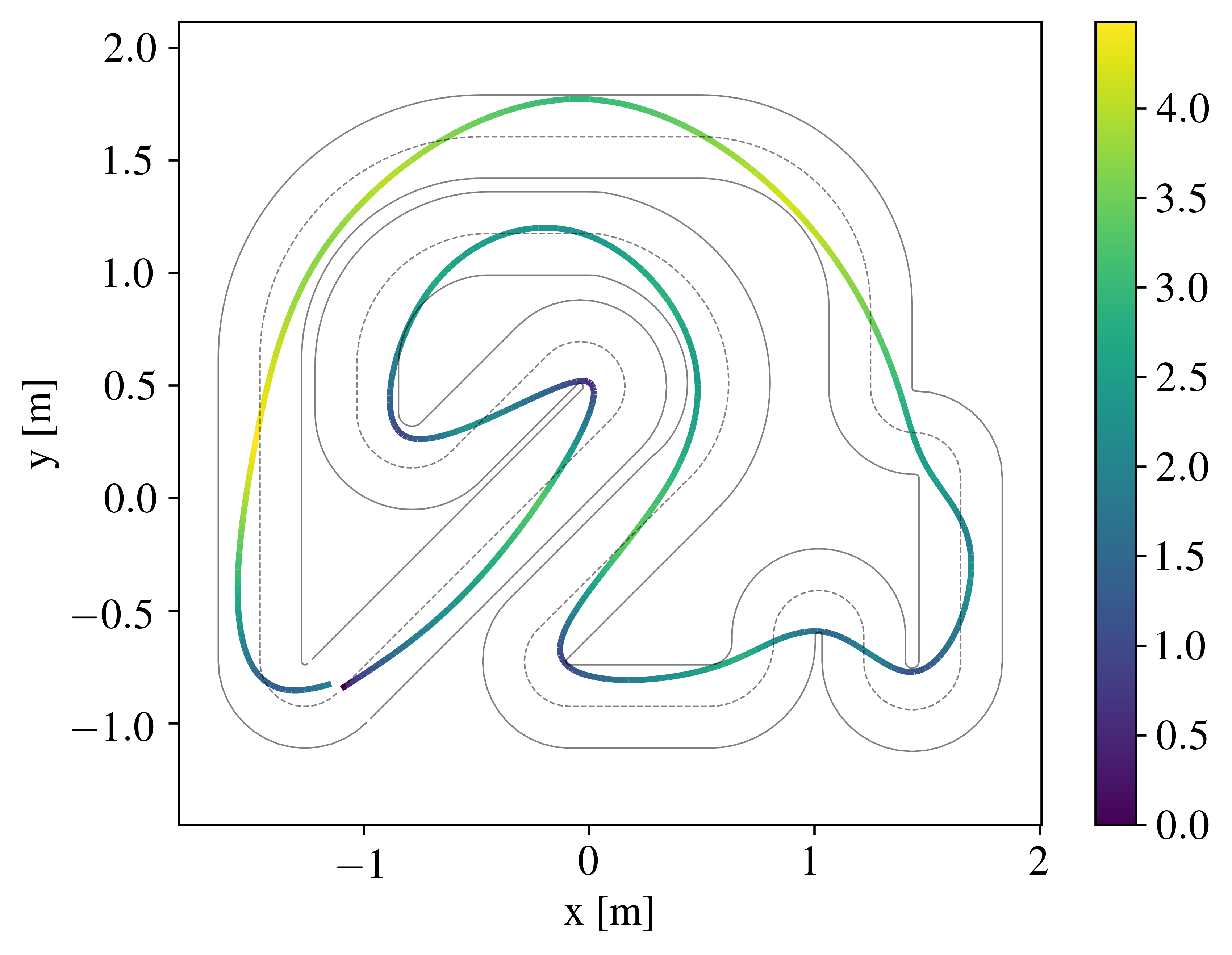}
	\caption{The racing line for an autonomous racing track at ETH Z\"urich. Color denotes speed in m/s.}
	\label{F:raclinglineexample}
\end{figure}

To this end, this paper makes the following contributions.
We propose a fully data-driven and computationally efficient algorithm to compute the racing line using Bayesian optimization.
Given (1) the xy-coordinates of the waypoints on the center line, (2) the track width, and (3) three vehicle parameters that can be physically measured, the algorithm computes the racing line in a few seconds.
It does not require closed-form expression or a parametric representation of the center line.
Teams participating in autonomous racing competitions can use this algorithm with ease to quickly precompute the racing line for a new track.
We derive racing lines for different tracks used for autonomous racing with 1/43 scale miniature cars at ETH Z\"urich \cite{Liniger2015} (Figure~\ref{F:raclinglineexample}) and 1/10 scale cars at UC Berkeley \cite{Rosolia2019}.
We also compare our approach against a baseline based on a random search.

\section{RACING LINE OPTIMIZATION}
\label{S:problem}

The objective is to determine a trajectory that requires minimum time to traverse a track for known vehicle dynamics.
We represent this dynamics by \(\dot{\mathbf{x}} = f_c\left(\mathbf{x}(t), \mathbf{u}(t)\right)\), where \(\mathbf{x}\) denotes the state of the vehicle and \(\mathbf{u}\) the set of control inputs.
Formally, the problem can be stated as
\begin{align}
& \minimize_{T, \mathbf{u}(t)} \ \ \ \ \int_{0}^{T} 1dt  \label{E:raceline}\\
& \begin{aligned}
\st \ \ \ \ 
& \dot{\mathbf{x}} = f_c\left(\mathbf{x}(t), \mathbf{u}(t)\right), \nonumber \\
& \mathbf{x}(0) = \mathbf{x}_{S} , \ \mathbf{x}(T) = \mathcal{X}_{F}, \nonumber \\
& \mathbf{x}(t) \in \mathcal{X}, \ \mathbf{u}(t) \in \mathcal{U}. \nonumber
\end{aligned}
\end{align}
Here, the second set of constraints includes an initial condition for the start line and a terminal condition for crossing the finish line.
\(\mathcal{X}\) and \(\mathcal{U}\) capture track and actuation constraints, respectively.
In discrete time, with vehicle dynamics given by \(\mathbf{x}_{k+1} = f_d(\mathbf{x}_k, \mathbf{u}_k)\), we can now formulate \eqref{E:raceline} as a finite horizon optimal control problem
\begin{align}
& \minimize_{T, \mathbf{u}_0, \mathbf{u}_1, \dots, \mathbf{u}_{T-1}} \ \ \ \ \sum_{0}^{T-1}1 \label{E:racelineFHC}\\
& \begin{aligned}
\st \ \ \ \ 
& \mathbf{x}_{k+1} = f_d(\mathbf{x}_k, \mathbf{u}_k), \nonumber \\
& \mathbf{x}_{0} = \mathbf{x}_{S} , \ \mathbf{x}_{T} = \mathcal{X}_{F}, \nonumber \\
& \mathbf{x}_k \in \mathcal{X}, \ \mathbf{u}_k \in \mathcal{U}, \nonumber \\
& \forall  k \in \left\lbrace 0, 1, \dots, T\right\rbrace. \nonumber
\end{aligned}
\end{align}
For more details, we refer the reader to \cite{Rosolia2019}.

Problem \eqref{E:racelineFHC} is an example of minimum time optimal control problem and is computationally hard to solve, especially in the presence of nonlinear constraints \cite{Athans2013}.
A standard method to solve \eqref{E:racelineFHC} is using dynamic programming (DP)\cite{Bertsekas2000}.
However, DP suffers from the curse of dimensionality.
It is computationally hard as the memory required increases exponentially with the number of states.
An iterative procedure that uses data from previous laps to reformulate \eqref{E:racelineFHC} with an updated terminal set and terminal cost is proposed in \cite{Rosolia2019}.
This method uses nonlinear optimization and is computationally more tractable.
Another interesting way to solve \eqref{E:racelineFHC}  is by using random sampling.
One can sample a feasible set of smooth trajectories between the start line and the finish line, and then evaluate minimum time to traverse each.
The random sampling method is inefficient because it requires a search over infinite feasible trajectories.
In this paper, we describe a new method where we guide the sampling of new trajectories using Bayesian optimization.

\section{PRELIMINARIES}
\label{S:prelim}

Before explaining our main algorithm, in this section, we briefly introduce the modeling of a Gaussian process (GP) and its use in Bayesian optimization (BayesOpt).

\subsection{Gaussian process regression}
\label{SS:gp}
A Gaussian process is a collection of random variables, any finite number of which have a joint Gaussian distribution.
Consider noisy observations \(x\) of an underlying function \(f: \mathbb{R}^n \mapsto \mathbb{R}\) through a Gaussian noise model: \(y = f(x) + \GaussianDist{0}{\sigma_n^2}\), \(x \in \mathbb{R}^n\).
A GP of \(y\) is fully specified by its mean function \(\mu(x)\) and covariance function \(k(x,x')\),
\begin{align}
\label{E:gp:prior}
\mu(x; \theta) &= \mathbb{E} [f(x)] \\
k(x,x'; \theta) &= \mathbb{E} [(f(x)\!-\!\mu(x)) (f(x') \!-\! \mu(x'))] + \sigma_n^2 \delta(x,x') \nonumber
\end{align}
where \(\delta(x,x')\) is the Kronecker delta function.
The hyperparameter vector \(\theta\) parameterizes the mean and covariance functions.
This GP is denoted by \(y \sim \mathcal{GP}(x)\).

Given the regression vectors \(X = [x_1, \dots, x_N]^T\) and the corresponding observed outputs \(Y = [y_1, \dots, y_N]^T\), we define training data by $\mathcal{D} = (X, Y)$. The distribution of the output \(y_\star\) corresponding to a new input vector \(x_\star\) is a Gaussian distribution \(\GaussianDist{\bar{y}_\star}{\sigma_\star^2}\), with mean and variance given by
\begin{subequations}
	\label{E:gp-regression}
	\begin{align}
	\bar{y}_\star &= \mu(x_\star) + K_\star K^{-1} (Y - \mu(X)) \label{E:gpmean}\\
	\sigma_\star^2 &= K_{\star \star} - K_\star K^{-1} K_\star^T \text, \label{E:gpvar}
	\end{align}
\end{subequations}
where \(K_\star = [k(x_\star, x_1), \dots, k(x_\star, x_N)]\), \(K_{\star \star} = k(x_\star, x_\star)\), and $K$ is the covariance matrix with elements \(K_{ij} = k(x_i, x_j)\).
The mean and covariance functions are parameterized by the hyperparameters $\theta$, which can be learned by maximizing the likelihood: \(\argmax_\theta \Pr(Y \vert X, \theta)\).
The covariance function \(k(x,x')\) indicates how correlated the outputs are at \(x\) and \(x'\), with the intuition that the output at an input is influenced more by the outputs of nearby inputs in the training data $\mathcal{D}$.
An elaborate description on modeling of GPs can be found in \cite{Rasmussen2006}.

\subsection{Bayesian optimization}
\label{SS:bayesopt}

Consider an unknown function \(f\) where we can only observe \(f(x)\) for a given \(x\).
Bayesian optimization focuses on maximizing (or minimizing) such a black-box function \(f\) over a feasible set \(\mathcal{X}\)
\begin{align*}
	\maximize & \ \ f(x) \\
	\st & \ \ x \in \mathcal{X}.
\end{align*}
Since we do not observe derivatives, first-order and second-order optimization methods cannot be used \cite{Frazier2018}.

BayesOpt learns a surrogate model of \(f\) using Gaussian process regression and sequentially updates the GP model as new data are observed.
It exploits two properties of GPs -- (1) GPs provide an estimate of uncertainty or confidence in the predictions through the predicted variance, and (2) GPs work well with small data sets.
We define an acquisition function \(\alpha\) that exploits the uncertainty in predictions to guide the search for optimal \(x\) by trading-off between exploration and exploitation.
Common choices for an acquisition function include expected improvement (EI) \cite{Jones1998} and noisy expected improvement (NEI) \cite{Letham2019}.
Thus, to search for the next sample to be evaluated, BayesOpt seeks to solve the following optimization problem sequentially
\begin{align}
\maximize_{x_\star} & \ \ \alpha\left(\bar{y}_\star(x_\star), \sigma_\star^2(x_\star)\right) \label{E:bayesoptex}\\
\st & \ \ x_\star \in \mathcal{X}, \nonumber
\end{align}
where \(\bar{y}_\star(x_\star)\) and \(\sigma_\star^2(x_\star)\) are defined in \eqref{E:gpmean} and \eqref{E:gpvar}, respectively.
We observe \(f(x_\star)\), update the GP model using new observation \(\left(x_\star,f(x_\star)\right)\), and problem \eqref{E:bayesoptex} is solved again.

BayesOpt is known for data-efficiency and is widely used in diverse applications such as tuning hyperparameters of complex deep neural networks \cite{Snoek2012}, learning data-efficient reinforcement learning (RL) policies for robotic manipulation tasks \cite{Englert2016}, tuning controller parameters in robotics \cite{Marco2016}, optimal experiment design for designing functional tests in buildings \cite{Jain2018} and recommender systems \cite{Li2010}.
For more details on BayesOpt, see \cite{Shahriari2015,Frazier2018}.
\section{ALGORITHM}
\label{S:algorithm}

In this section, we describe our main algorithm for computing the racing line using Bayesian optimization.
We perform the following three steps.
First, we parameterize a trajectory using an \(n\)-dimensional vector (\(n\) being the number of waypoints) that fully characterizes a smooth trajectory on the racing track.
This parameterization allows us to randomly sample feasible and smooth candidate trajectories from the start line to the finish line.
Second, we evaluate the minimum time to traverse these parameterized trajectories while driving the vehicle at the limits of friction following the approach in \cite{Lipp2014}.
This allows us to assess the quality or fitness of any parameterized trajectory in terms of minimum lap time.
Lastly, we learn a GP model that is trained on sampled trajectories (\(n\)-dimensional vector) as input and minimum time to traverse these trajectories as output.
The model is initialized with randomly sampled trajectories.
Following which the sampling is guided using Bayesian optimization to iteratively search for a trajectory that can potentially further reduce the lap time.
In the following subsections, we explain each of the above steps in detail.
The code is available at https://github.com/jainachin/bayesrace.

\subsection{Parameterization}
\label{SS:param}

\begin{figure}[t!]
	\centering
	\includegraphics[width=1\columnwidth]{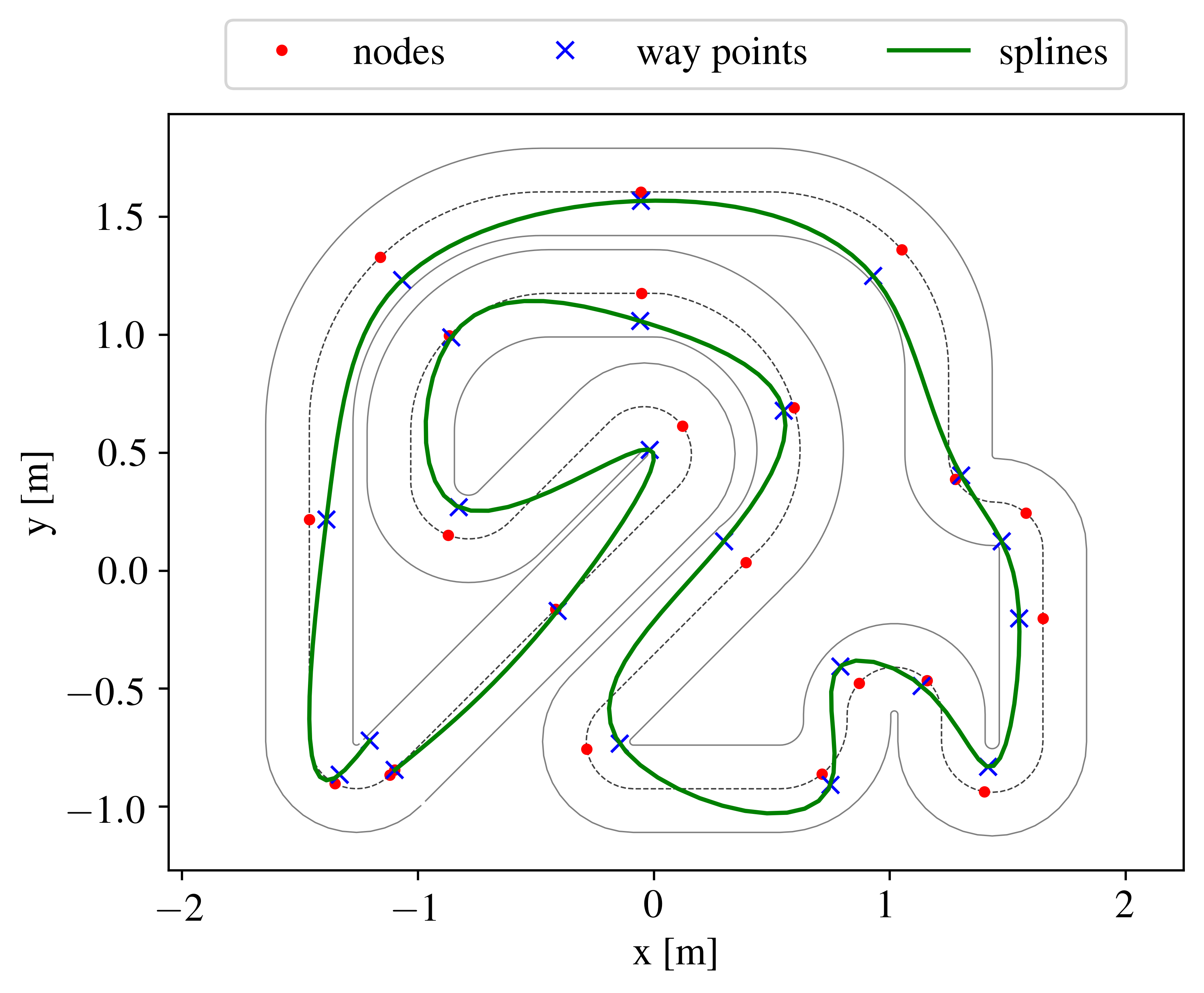}
	\caption{An example of a randomly sampled trajectory obtained after parameterization given by deviation from the center line.}
	\label{F:random}
\end{figure}

For a given track, we assume that we know the center line, specifically the xy-coordinates of the waypoints on the center line, and the track width (which can be constant or variable along the center line).
We begin with defining nodes along the center line.
These are depicted with red markers in Figure~\ref{F:random}.
The number of nodes (same as variable \(n\) above) depends upon the length of the track.
We select more nodes near the corners to prevent cutting around them.
Next, we define waypoints by perturbing \(i^{th}\) node by \(w_i\) in the lateral direction (normal to the center line).
Thus, the parameterization of the track is given by \(\mathbf{w}:=[w_1, w_2, \dots, w_n]\), where each \(w_i\) can vary between \([-\frac{w_T}{2}, \frac{w_T}{2}]\), \(w_i=0\) corresponds to the center line and \(w_T\) is the width of the track.
These waypoints sampled uniformly in the range \([-\frac{w_T}{2}, \frac{w_T}{2}]\) are shown as blue markers in Figure~\ref{F:random}.
The dimensionality of \(\mathbf{w}\) affects the convergence rate of Bayesian optimization in Section~\ref{SS:raceopt}.
Thus, it is advisable to choose less than 30 nodes.
Note that if xy-coordinates of the waypoints are used for parameterization, we will have twice as many parameters as we need one parameter for each \(x_i\) and \(y_i\) for all \(n\) nodes.
In our parameterization, we exploit the fact that we know the center line.
Moving \(w_i\) in the direction normal to it gives the xy-coordinates \((x_i,y_i)\) of the \(i^{th}\) waypoint.
Finally, to generate a smooth trajectory, the waypoints are joined by 2D cubic spline interpolation as shown in green in Figure~\ref{F:random}.

\subsection{Minimum time to traverse on a fixed trajectory}
\label{SS:mintime}

Our goal is to evaluate the fitness of a candidate trajectory like the one randomly sampled in Figure~\ref{F:random}.
To calculate the minimum time to traverse a fixed trajectory, we use a friction circle model with a rear-wheel drive given by
\begin{align}
	m\begin{bmatrix} \ddot{x}\\\ddot{y} \end{bmatrix} = \begin{bmatrix} \cos(\phi)  & -\sin(\phi)\\\sin(\phi)  & \cos(\phi) \end{bmatrix} \begin{bmatrix} F_{\mathrm{long}}\\F_{\mathrm{lat}}\end{bmatrix},
	\label{E:frictioncircle}
\end{align}
where \(m\) is the mass of the vehicle and \(\phi\) the orientation of the vehicle defined as a function of position \((x,y)\) in the global frame.
The inputs to the model are a force in the longitudinal direction \(F_{\mathrm{long}}\) and a force in the lateral direction \(F_{\mathrm{lat}}\) defined in the frame attached to the vehicle.
We enforce a constraint for the friction circle
\begin{align}
	\sqrt{F_{\mathrm{long}}^2 + F_{\mathrm{lat}}^2} & \leq \mu_s m g,
	\label{E:fcons1}
\end{align}
where \(\mu_s\) is the static coefficient of friction, \(g\) is the acceleration due to gravity, and a constraint for the maximum possible force with the rear-wheel drive
\begin{align}
	F_{\mathrm{long}} & \leq \frac{l_f}{l_f+lr}~\mu_s m g,
	\label{E:fcons2}
\end{align}
where \(l_f\) and \(l_r\) are the distance of the center of gravity from the front and the rear wheels in the longitudinal direction, respectively.
The model ignores the effect of tire slips.
The advantage of using the friction circle model is that it requires minimum effort in system identification with only three parameters to identify, namely \(m\), \(l_f\) and \(l_r\).
The kinematic bicycle model \cite{Kong2015} also requires the same parameters.
The true behavior of the car is represented more closely by the dynamic bicycle model \cite{Kong2015}, especially during high-speed cornering, which also includes forces due to tire slips.
However, it is also much harder to tune as it has many more parameters.
We are currently designing a learning-based control algorithm where the controller uses a simple model like the friction circle (and hence the knowledge of the optimal racing line as proposed in this paper) and then iteratively learns the unmodeled dynamics from data.

For a fixed trajectory, calculation of minimum time to traverse and the corresponding speed profile will require solving \eqref{E:raceline}, where the vehicle dynamics is given by \eqref{E:frictioncircle}, with an additional constraint that \((x,y)\) must lie on the trajectory.
It turns out this problem is much easier to solve.
By transforming the problem from a generalized position space to a path coordinate space and subsequently applying the nonlinear change of variables, the problem of calculating minimum time over a fixed path can be formulated as a convex optimization problem \cite{Verscheure2009}.
For the friction circle model \eqref{E:frictioncircle} with additional constraints \eqref{E:fcons1} and \eqref{E:fcons2}, the optimization is still convex \cite{Lipp2014}.
Now, given a trajectory parameterized by \(\mathbf{w}\), since the number of waypoints is chosen to be small by choice, we re-sample 100 waypoints after fitting cubic splines and then apply the result from \cite{Lipp2014} to calculate the minimum time to traverse.
For our experiments on the chosen tracks, 100 waypoints were sufficient. 
For longer tracks, we recommend re-sampling more waypoints.
The steps are summarized in Algorithm~\ref{A:mintime}.
\begin{algorithm}[t]
	\caption{Minimum time to traverse on a fixed trajectory}
	\label{A:mintime}
	\begin{algorithmic}[1]
		\Procedure{CalcMinTime}{$\mathbf{w}$}
		\State get \((x_i, y_i)\) from \(w_i \ \forall i \in \{1,2,\dots,n\}\)
		\State fit cubic splines on the waypoints given by \((x_i, y_i) \)
		\State re-sample way points with finer discretization \((\hat{x}_k, \hat{y}_k)\)
		\State \textit{return} minimum time to traverse on \((\hat{x}_k, \hat{y}_k)\) using \cite{Lipp2014}
		\EndProcedure
	\end{algorithmic}
\end{algorithm}

\subsection{Guiding sampling using Bayesian optimization}
\label{SS:raceopt}

The central idea here is to use the uncertainty estimate in the predictions of a GP model to guide how the \(w_i\)'s should be changed to reduce lap times.

To initialize a GP model, we randomly sample parameters \(\mathbf{w}_j \ \forall j \in \{1,2,\cdots,10\}\)  to generate 10 trajectories like the one shown in Figure~\ref{F:random}.
We then evaluate minimum time to traverse each trajectory \(\tau_j \ \forall j \in \{1,2,\cdots,10\}\) using Algorithm~\ref{A:mintime}.
The parameters of the trajectory \(\mathbf{w}\) are used as inputs and the minimum lap time \(\tau\) as output to define a  GP model
\begin{align}
	\tau \sim \mathcal{GP}(\mathbf{w}) := \mathcal{N}\left(\bar{\tau}, \sigma_{\tau}^2\right).
	\label{E:gpinit}
\end{align}
The output of the GP model \(\tau\) is a normal distribution whose mean \(\tau\) and variance \(\sigma_{\tau}^2\) are given by \eqref{E:gpmean} and \eqref{E:gpvar}, respectively.

Recall, our objective is to determine a trajectory that minimizes the lap time with given vehicle dynamics.
At this stage, even the best trajectory, whose index is given by \(\argmin_{\{1,2,\dots,10\}} \tau_j\), is far from the optimal racing line.
We apply Bayesian optimization with expected improvement as the acquisition function to determine the next candidate trajectory that would potentially reduce the lap time further by solving the following optimization problem
\begin{align}
& \maximize_{\mathbf{w}} \ \ \ \ \mathbb{E}\left(\left[\tau_{\mathrm{best}} - \mathcal{GP}(\mathbf{w})\right]^+\right) \label{E:bayesopt}\\
& \begin{aligned}
\st \ \ \ \ 
& -\frac{w_T}{2} \leq w_i \leq \frac{w_T}{2} \  \  \forall i \in \{1,2,\dots,n\},  \nonumber\\
\end{aligned}
\end{align}
where \(\tau_{\mathrm{best}}\) is the minimum lap time observed so far and \(\left[x\right]^+ := \mathrm{max}(0,x)\).
The optimal solution of \eqref{E:bayesopt} denoted by \(\mathbf{w}^\star\) is evaluated using Algorithm~\ref{A:mintime}.
Denote the outcome by \(\tau^\star\).
The GP model in \eqref{E:gpinit} is updated using this new observation \((\mathbf{w}^\star,\tau^\star)\), and the optimization problem \eqref{E:bayesopt} is solved iteratively until convergence.
This procedure to determine the optimal racing line is summarized in Algorithm~\ref{A:bayesopt}.
The algorithm converges in a finite number of iterations with the racing line and the sequence of control inputs to drive the racing line.
In Section~\ref{S:experiments}, we also run experiments with a different acquisition function -- noisy expected improvement.
For details on how to define the cost in the optimization problem \eqref{E:bayesopt} in this case, see \cite{Letham2019}.
\begin{algorithm}[t]
	\caption{Racing line using Bayesian optimization}
	\label{A:bayesopt}
	\begin{algorithmic}[1]
		\Procedure{Initialization}{}
		\For{\(j \in \{1,2,\dots,10\}\)}
		\State randomly sample a new trajectory parametrized by \(\mathbf{w}_j\)
		\State compute min time to traverse \(\tau_j\) using Algorithm~\ref{A:mintime}
		\EndFor
		\State initialize training data \(\mathcal{D}:=\bigcup_{j=1}^{10} (\mathbf{w}_j,\tau_j)\)
		\State learn a GP model \(\tau \sim \mathcal{GP}(\mathbf{w})\)
		\EndProcedure
		\Procedure{Bayesian Optimization}{}
		\While{lap time not converged}
		\State determine candidate trajectory \(\mathbf{w}^\star\) by solving \eqref{E:bayesopt}
		\State compute min time to traverse \(\tau^\star\) using Algorithm~\ref{A:mintime}
		\State add new sample to training data \(\mathcal{D} = \mathcal{D}~\bigcup~(\mathbf{w}^\star, \tau^\star)\)
		\State update the \(\mathcal{GP}\) model using \(\mathcal{D}\)
		\EndWhile
		\State \textit{return}  \(\mathbf{w}^\star\) and corresponding way points \((x_i, y_i)\)
		\EndProcedure
	\end{algorithmic}
\end{algorithm}
\section{EXPERIMENTS}
\label{S:experiments}

\begin{figure*}[t!]
	\centering
	\includegraphics[width=0.94\columnwidth]{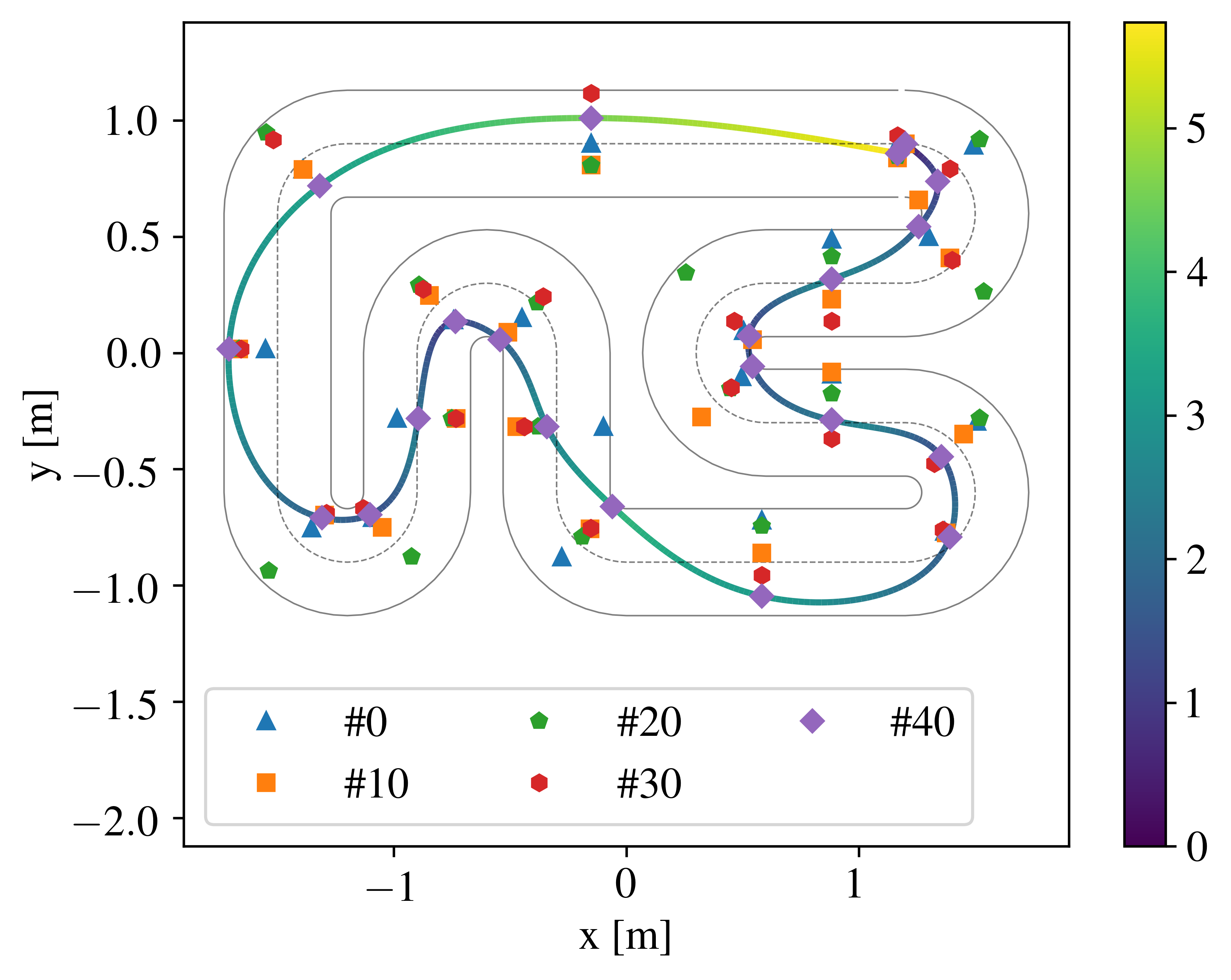}
	\includegraphics[width=1\columnwidth]{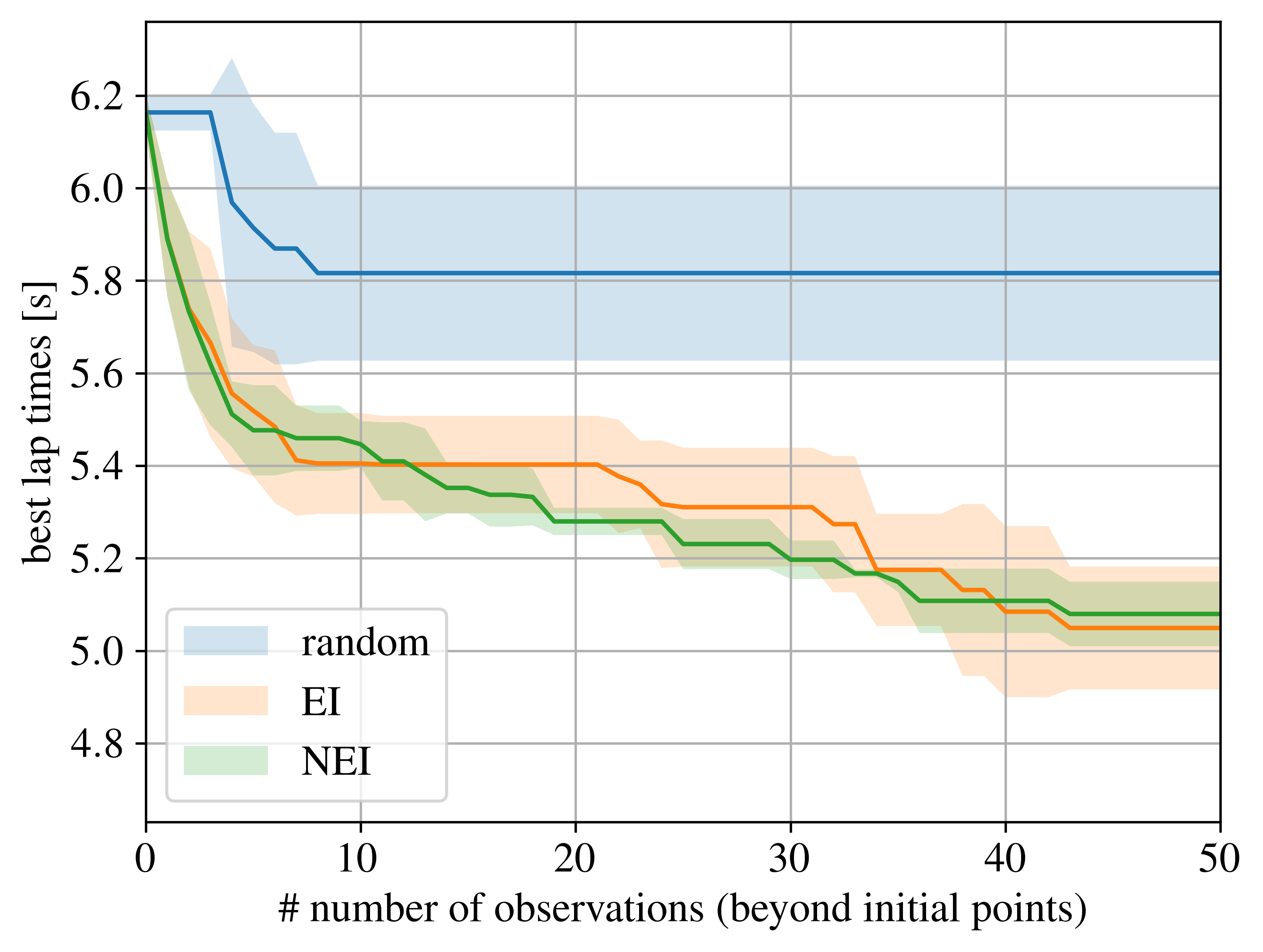}
	\caption{ETHZ1: Track located at Automatic Control Lab at ETH Z\"urich. The direction of racing is clockwise.}
	\vspace{10pt}
	\label{F:ETHZ1}
\end{figure*}

\begin{figure*}[t!]
	\centering
	\includegraphics[width=0.98\columnwidth]{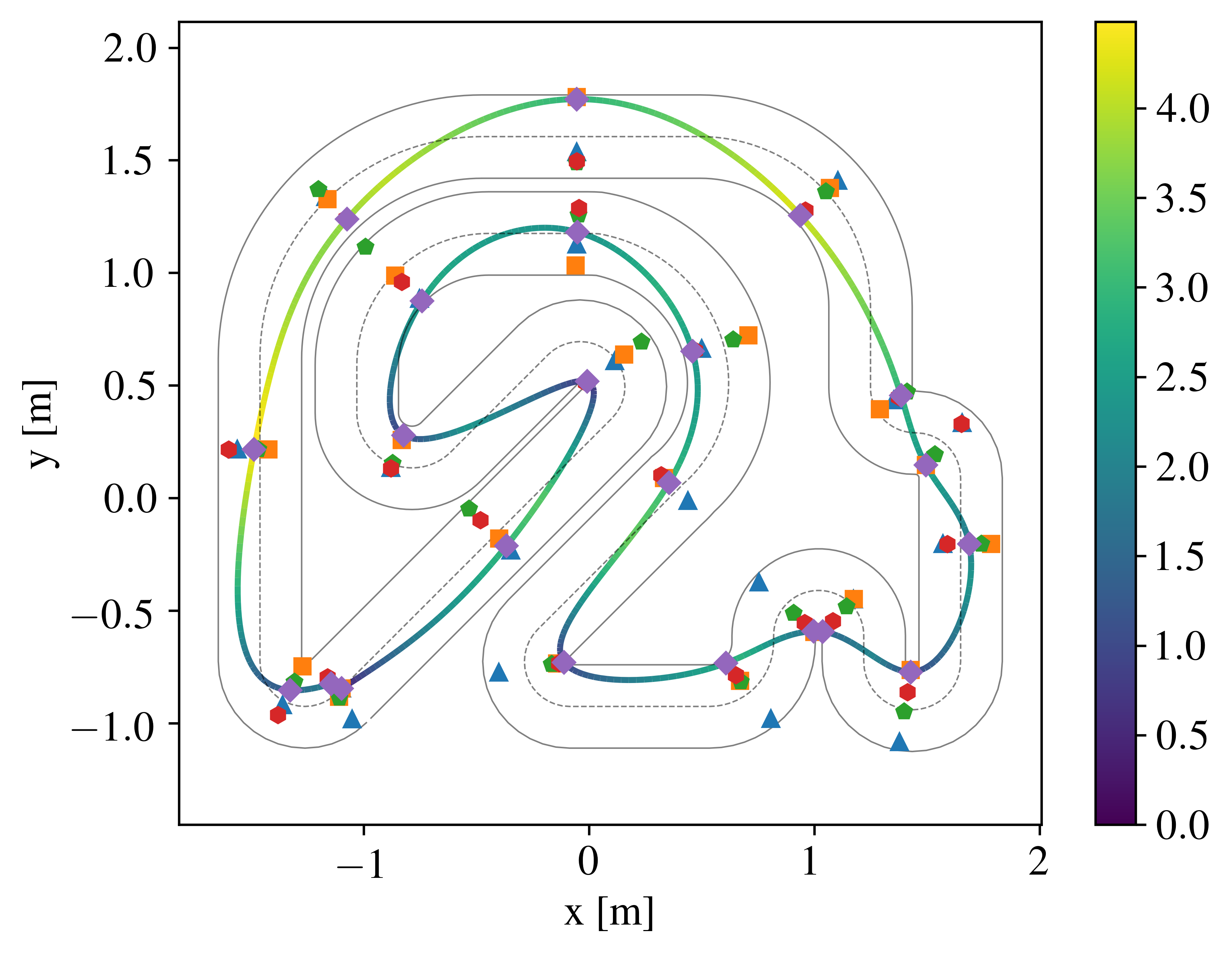}
	\includegraphics[width=1\columnwidth]{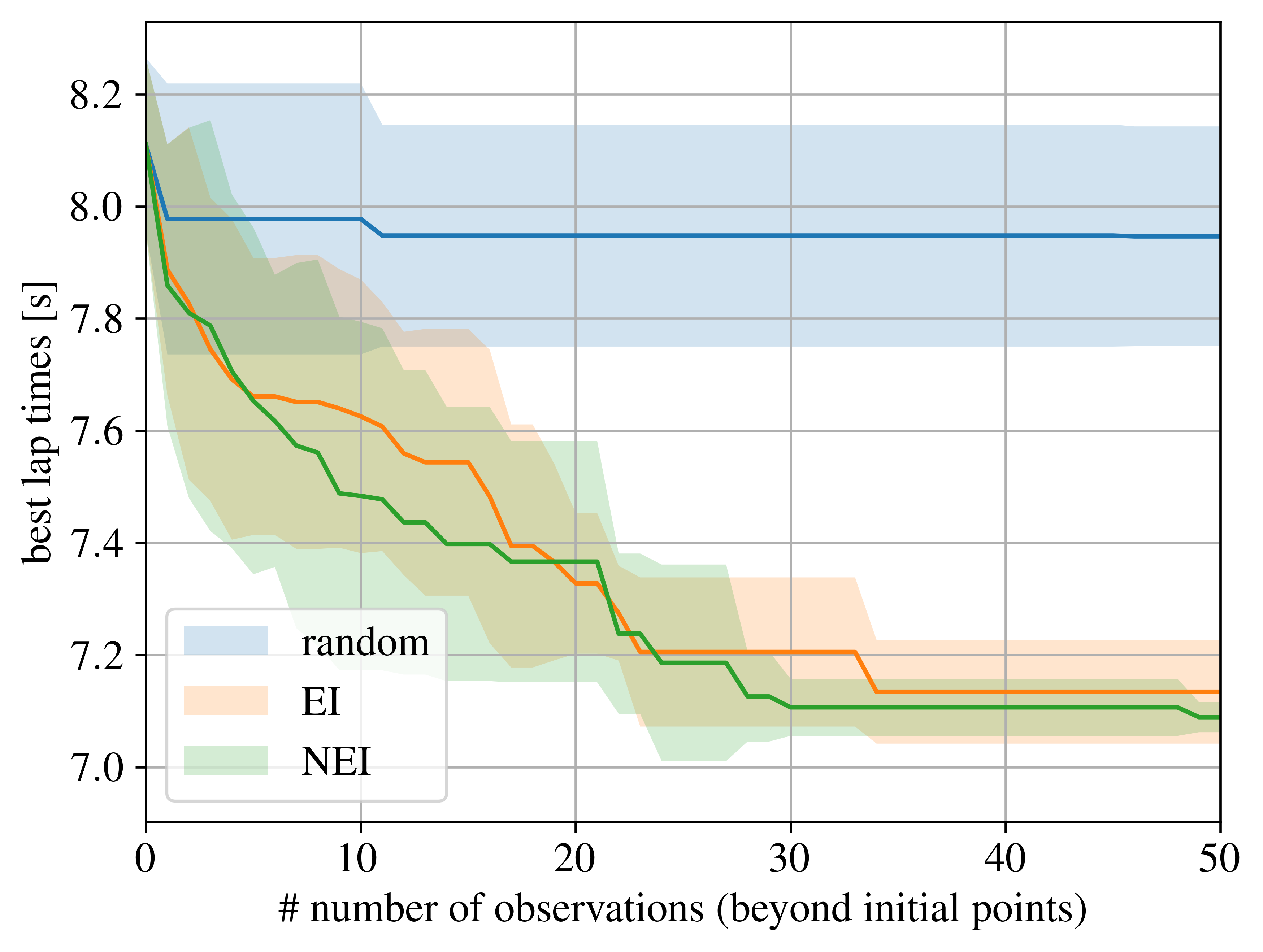}
	\caption{ETHZ2: Track located at the department of Mechanical Engineering at ETH Z\"urich. The direction of racing is anti-clockwise.}
	\vspace{10pt}
	\label{F:ETHZ2}
\end{figure*}

\begin{figure*}[t!]
	\centering
	\includegraphics[width=0.90\columnwidth]{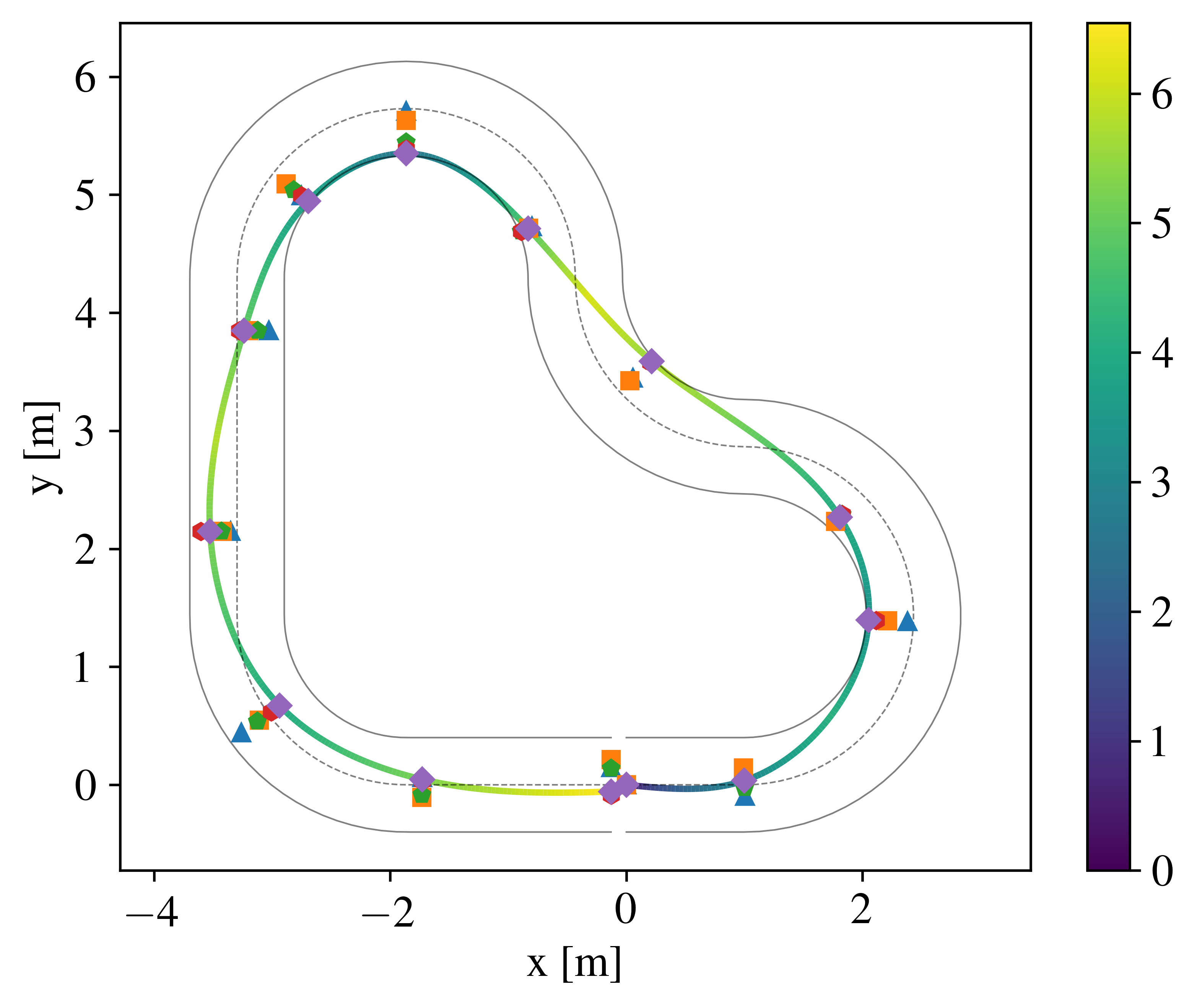}
	\includegraphics[width=1\columnwidth]{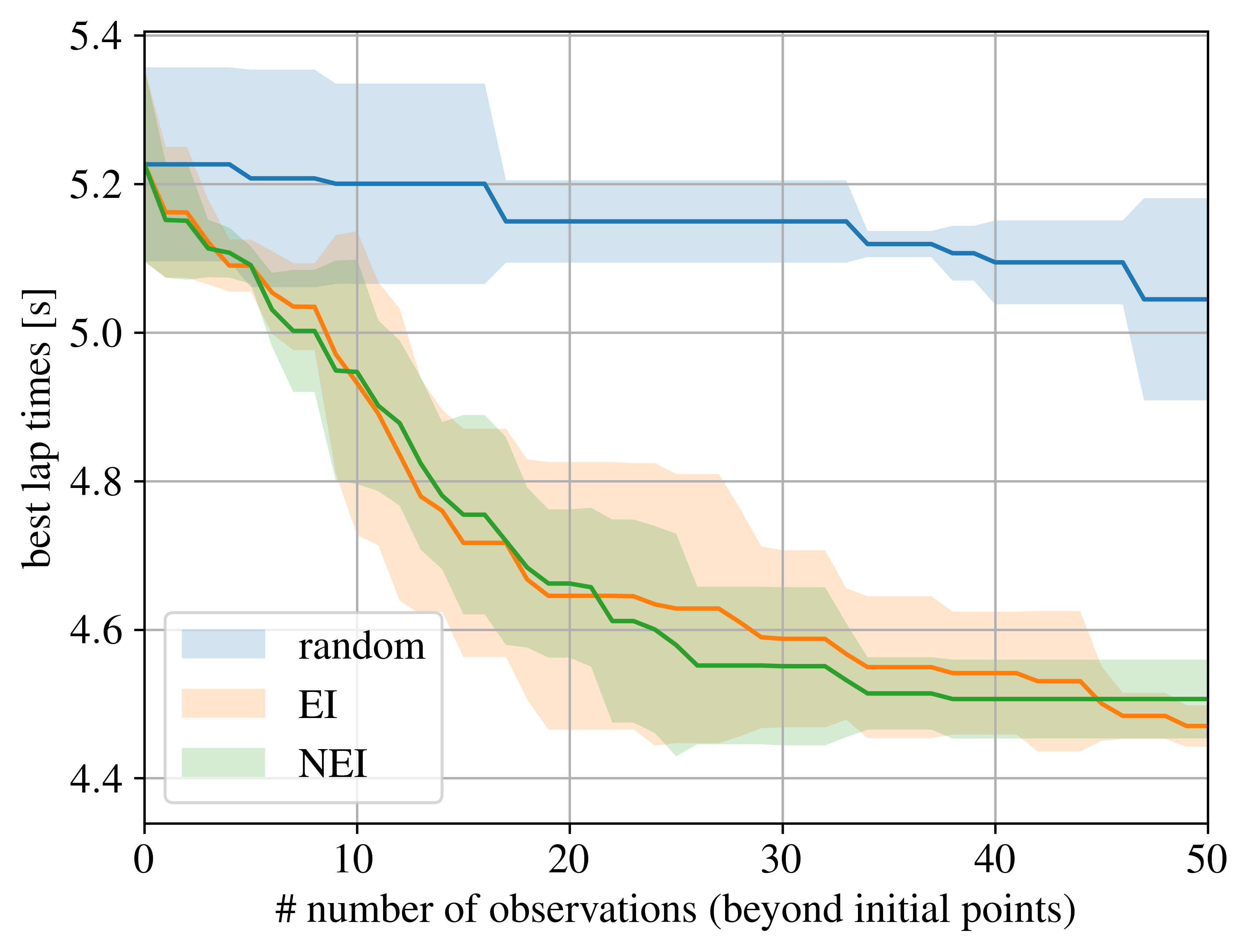}
	\caption{UCB: Track located at Model Predictive Control Lab at UC Berkeley. The direction of racing is anti-clockwise.}
	\label{F:UCB}
\end{figure*}

We compute the racing lines for two tracks at ETH Z\"urich used for autonomous racing with 1/43 scale cars \cite{Liniger2015}, shown in Figure~\ref{F:ETHZ1} and Figure~\ref{F:ETHZ2}, and a track at UC Berkeley used with 1/10 scale cars \cite{Rosolia2019}, shown in Figure~\ref{F:UCB}.


It is assumed that, in all three cases, the cars start at the marked location on the tracks with zero initial speed.
The GP models are initialized by sampling 10 randomly generated trajectories

We compare three methods for sampling new trajectories: (1) uniform random sampling, (2) BayesOpt with EI acquisition function, and (3) BayesOpt with NEI acquisition function.
We keep a record of the best lap time as more trajectories are sampled.
For each track, the decrease in the best lap time with each method is shown on the right in Figure~\ref{F:ETHZ1}-\ref{F:UCB}.
We observe BayesOpt converges to good racing lines in less than 50 new observations while the uniform random sampling is highly sample inefficient.
We also show 95\% confidence bounds for convergence obtained by running each method multiple times.
Computing these racing lines requires less than three minutes using CVXPY \cite{CVXPY} for Algorithm~\ref{A:mintime} and BoTorch \cite{Balandat2019} for Algorithm~\ref{A:bayesopt}.
Algorithm~\ref{A:mintime} requires more than 80\% of total compute time.
Our current implementation of Algorithm~\ref{A:mintime} can be made 10x more efficient by using code generation in C++ with FORCES Pro \cite{ForcesPro}.

\begin{figure}[t]
	\centering
	\includegraphics[width=1\columnwidth]{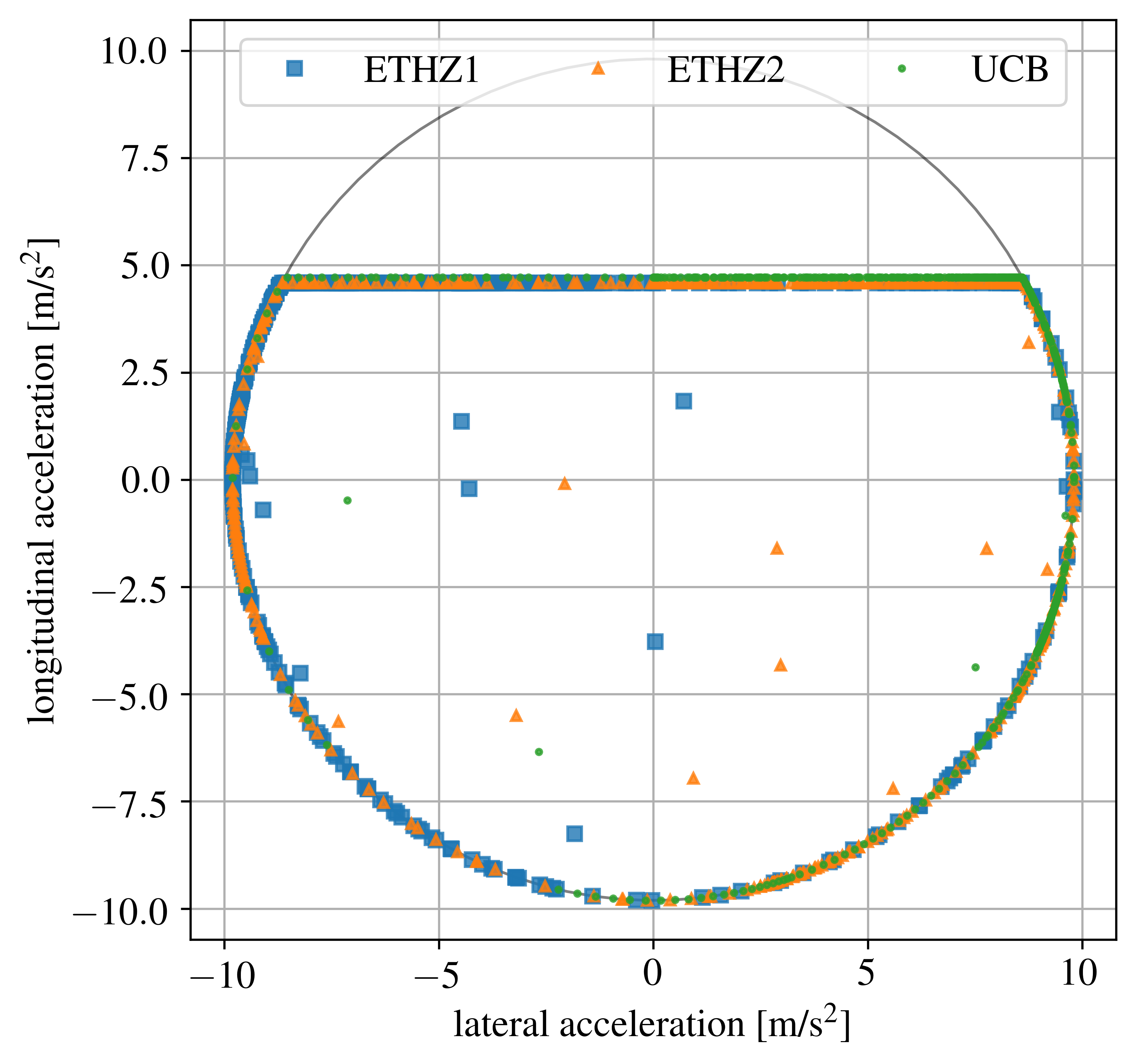}
	\caption{GG diagram: ETHZ tracks (ETHZ1 and ETHZ2) are used for racing 1/43 scale cars and UCB for 1/10 scale cars. Corresponding vehicle parameters are used.}
	\label{F:frictioncircle}
\end{figure}

In Figure~\ref{F:ETHZ1}-\ref{F:UCB}, on the left we demonstrate how each node is strategically moved in the lateral direction by BayesOpt to decrease lap times over iterations.
The nodes corresponding to the best lap after initialization are denoted by \markerthree, the best lap after 10 new observations by \markerfour,  the best lap after 20 new observations by \markerfive,  the best lap after 30 new observations by \markersix, and the best lap after 40 new observations by \markerdiamond.
The racing line is shown corresponding to \markerdiamond.
The longitudinal and lateral acceleration for all three tracks are shown on a GG diagram in Figure~\ref{F:frictioncircle}.
It is clear that at most times the vehicle is operating on the boundaries of the friction circle to minimize lap times.

\section{CONCLUSION}
\label{S:conclusion}

We introduce a fully data-driven method to compute the racing line using Bayesian optimization.
The algorithm only requires the xy-coordinates of the waypoints, the track width, and three vehicle parameters that can be physically measured -- mass and distance of the center of gravity from the front and rear wheels.
It is computationally efficient compared to standard methods like dynamic programming and random search and requires minimal manual effort.
We demonstrate the algorithm on three different tracks.
The teams participating in autonomous racing competitions can use our algorithm to quickly compute the racing line for a new track for the design of a motion planner and a controller.




\bibliographystyle{unsrt}
\bibliography{bayesrace}


\addtolength{\textheight}{-12cm}   


\end{document}